\documentclass{svproc}
\usepackage{graphicx}
\usepackage{url}

\usepackage{multirow}
\newcommand{\repeatthanks}{\textsuperscript{\thefootnote}}
\begin{document}
\mainmatter              
\title{Evaluating Deep Learning Approaches for Covid19 Fake News Detection}
\titlerunning{Covid19 Fake News Detection}  
%
\author{Apurva Wani\inst{1}\thanks{Authors contributed equally} \and Isha Joshi\inst{1}\repeatthanks \and  Snehal Khandve\inst{1}\repeatthanks \and Vedangi Wagh\inst{1}\repeatthanks \and \\ Raviraj Joshi\inst{2}} 
\authorrunning{Apurva Wani et al.} 
%
%
\institute{Pune Institute of Computer Technology, Pune
\and
Indian Institute of Technology Madras, Chennai\\
\email{{apurva.wani06, ishajoshi.211, snehal.khandve07, vedangikwagh, ravirajoshi}@gmail.com}}

\maketitle              

\begin{abstract}
Social media platforms like Facebook, Twitter, and Instagram have enabled connection and communication on a large scale. It has revolutionized the rate at which information is shared and enhanced its reach. However, another side of the coin dictates an alarming story. These platforms have led to an increase in the creation and spread of fake news. The fake news has not only influenced people in the wrong direction but also claimed human lives. During these critical times of the Covid19 pandemic, it is easy to mislead people and make them believe in fatal information. Therefore it is important to curb fake news at source and prevent it from spreading to a larger audience. We look at automated techniques for fake news detection from a data mining perspective.  We evaluate different supervised text classification algorithms on Contraint@AAAI 2021 Covid-19 Fake news detection dataset. The classification algorithms are based on Convolutional Neural Networks (CNN), Long Short Term Memory (LSTM), and Bidirectional Encoder Representations from Transformers (BERT). We also evaluate the importance of unsupervised learning in the form of language model pre-training and distributed word representations using unlabelled covid tweets corpus. We report the best accuracy of 98.41\% on the Covid-19 Fake news detection dataset.
\end{abstract}

\section{Introduction}
Technology has been dominating our lives for the past few decades. It has changed the way we communicate and share information. The sharing of information is no longer constrained by physical boundaries. It is easy to share information across the globe in the form of text, audio, and video. An integral part of this capability is the social media platforms. These platforms help in sharing personal opinions and information with much a wider audience. They have taken over traditional media platforms because of speed and focussed content. However, it has become equivalently easy for nefarious people with malicious intent to spread fake news on social media platforms. 

Fake news is defined as a verifiably false piece of information shared intentionally to mislead the readers \cite{shu2017fake}. It has been used to create a political, social, and economic bias in the minds of people for personal gains. It aims at exploiting and influencing people by creating fake content that sounds legit. On the extreme end, fake news has even led to cases of mob lynching and riots \cite{Fakenews79:online}. Thus, it is extremely important to stop the spread of fake content on internet platforms. It is especially desirable to control fake news during the ongoing Covid-19 crisis \cite{tasnim2020impact}. The pandemic has made it easy to manipulate a mentally stranded population eagerly waiting for this phase to end. Some people have reportedly committed suicide after being diagnosed with covid due to the misrepresentation of covid in social and even mainstream media \cite{Coronavi12:online}. The promotion of false practices will only aggravate the covid situation.

Recently, researchers have been actively working on the task of fake news detection. While manual detection \cite{PolitiFa30:online,BOOMCoro43:online,HomeNews39:online} is the most reliable method it has limitations in terms of speed. It is difficult to manually verify the large volumes of content generated on the internet. Therefore automatic detection of fake news has gained importance. Machine learning algorithms have been employed to analyze the content on social media for its authenticity \cite{wang2017liar}. These algorithms mostly rely on the content of the news. The user characteristics, the social network of the user, and the polarity of their content are another set of important signals \cite{zhang2020overview}. It is also common to analyze user behavior on social platforms and assign them a reliability score. The fake news peddlers might not exhibit normal sharing behavior and will also tend to share more extreme content. All these features taken together provide a more reliable estimate of authenticity.

In this work, we are specifically concerned with fake news detection related to covid. The paper describes systems evaluated for Contraint@AAAI 2021 Covid-19 Fake news detection shared task \cite{patwa2021overview}. The task aims in improving the classification of the news based on Covid-19 as fake or real. The dataset shared is created by collecting data from various social media sources such as Instagram, Facebook, Twitter, etc.

The fake news detection task is formulated as a text classification problem. We solely rely on the content of the news and ignore other important features like user characteristics, social circle, etc which might not always be available. We evaluate the recent advancements in deep learning based text classification algorithms for the task of fake news detection. The techniques include pre-trained models based on BERT and raw models based on CNN and LSTM. We also evaluate the effect of using monolingual corpus related to covid for language model pretraining and training word embeddings. In essence, we rely on these models to automatically capture discriminative linguistic, style, and polarity features from news text helpful for determining authenticity.

\section{Related Works}
Fake news detection on traditional outlets of news and articles solely depends on the reader's knowledge about the subject and the article content. But detection of fake news that has been transmitted via social media has various cues that could be taken into consideration. One of the cues can be finding a user's credibility by analyzing their followers, the number of followers, and their behavior as well as their registration details. In addition to these details \cite{castillo2011information} have used other factors such as attached URLs, social media post propagation features, and content-based hybrid model for classifying news as fake or genuine. Another research \cite{kwon2013prominent} based on structural properties of the social network is used for defining a “diffusion network” which is the spread of a particular topic. This diffusion network together with other social network features can be helpful in the classification of rumors in social media with classifiers like SVM, random forest, or decision tree.

Besides using the characteristics of user-patterns and user details who share fake news, another context useful for classifying any social media news post is the comments section. \cite{zhao2015enquiring} have performed a linguistic study and have found comments like “Really?”, “Is it true?” in the comment section of some of the fake posts. They have further implemented a system that clusters such inquiry phrases in addition to clustering simple phrases for classifying rumors. 

Another approach of considering the tri-relationship between the publishers, the news articles, and users of fake news can be considered. This relationship has been used to create a tri-relationship embedding framework TriFN in \cite{shu2019beyond} for detection of fake news articles on social media. Four types of embeddings namely the news content embedding, user embedding, user-news interaction embeddings as well as publisher-news relation embeddings with contributions to spread fake news are generated coupled with a semi-supervised classifier are used in TriFN to identify fake news. 

The propagation knowledge of fake news articles such as its path construction and transformation can also be useful for primary detection of fake news \cite{liu2018early}. Further, this transformation path has been represented into vectors for classification with deep neural network architectures namely the RNN for global variation and CNN for local variation of the path.

Apart from user-context and social-context features the content of the fake news has also been a proven way of detecting fake news and rumors. A recent approach utilizes explicit as well as latent features of the textual information for further classification of news \cite{zhang2020fakedetector}. Basic Deep Convolutional neural networks have also been used to get contextual information features from fake news articles for identifying them \cite{kaliyar2020fndnet}. 

\begin{figure}
    \centering
    \includegraphics[width=0.45\textwidth]{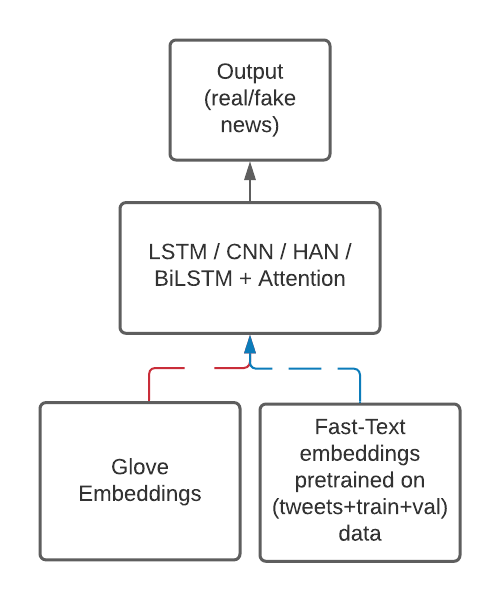}
    \includegraphics[width=0.4\textwidth]{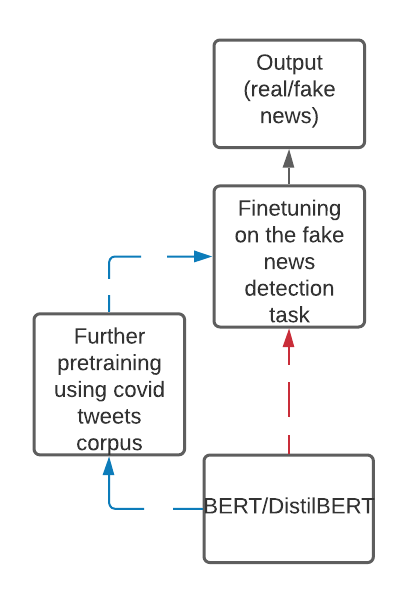}
    \caption{Model summary showing two approaches of Simple models(L) and two approaches of Transformer based models(R) using different colours.}
    \label{fig:my_label}
\end{figure}

\section{Architecture Details}
In this section, we describe the techniques we have used for text classification. We also describe the hyper-parameters used in each of these models. The model summary is shown in Fig. \ref{fig:my_label} for the two types of architectures explored in this work.
\subsection{CNN}
Although CNN is mostly used for image recognition tasks, text classification is also recognized as one of the applications of CNN \cite{kim2014convolutional}. The CNN layers extract useful features from the word embeddings to generate the output. The 300-dimensional fast text embeddings are used as input to the first layer. We use a slightly deep architecture with initial five parallel 1D Conv layers. The kernel size for these parallel convolutions is size 2,3,4,5,6. The number of filters used in these conv layers is 128. The output of these conv layers are concatenated and then fed to two sequential blocks of 1D conv layer followed by 1D MaxPooling layer. Three dense layers of sizes 1024, 512, and 2 are subsequently added to the entire architecture. There is a dropout of 0.5 added after the final two conv layers and the first two dense layers. This CNN model is trained on a batch size of 64 samples and an Adam optimizer is used. The batch size and optimizer are constant for all non-BERT models.
\subsection{LSTM}
Long Short-Term Memory (LSTM) is a type of Gated-RNN architecture along with the feedback connections \cite{hochreiter1997long}. With the input length equal to the length of the longest tweets in train data, the embedding layer is the first layer. It is followed by a single LSTM layer with 128 units, a single dropout layer with a dropout rate of 0.5, and two dense layers with units 128 and 2 respectively.
\subsection{Bi-LSTM + Attention}
The additional feature that the Bi-LSTM network offers is that it considers the input sequence from both the forward and reverse direction. This sequential model has a first embedding layer similar to the previous models. The next layer is a bidirectional LSTM with 256 units in each direction followed by an attention layer and two dense layers with 128 and 2 units. The structure of the attention layer is borrowed from \cite{zhou2016attention}.
\subsection{HAN}
Hierarchical Attention Networks (HAN) is based on LSTM and comprises of four sequential levels - word encoder, word-level attention, sentence encoder, and sentence-level attention \cite{yang2016hierarchical}. Each data sample is divided into a maximum of 40 sentences and each sentence consists of a maximum of 50 words. The word encoder is a bidirectional LSTM that works on word embeddings of individual sentences to produce hidden representation for each word. The word-level attention helps us to extract important words that contribute to the meaning of the sentence. These informative words which conceive the complete meaning of the sentence are aggregated to form sentence vectors. The sentence vectors are processed by another bidirectional LSTM referred to as sentence encoder. The sentence-level attention layer measures the importance of each sentence and sentences which provide the most significant information for classification are summarized to get a document vector that contains the gist of the entire data sample.
\subsection{Transformers}
Transformers have outperformed previous sequential models in various NLP tasks \cite{vaswani2017attention}. The major component of transformers is self-attention which is a variant of the attention mechanism. Self-attention is used for generating a contextual embedding of any given word in the input sentence with respect to other words in the sentence. The major advantage of transformers over RNNs \cite{sutskever2014sequence} was that it led to parallelization of the process which made it possible to take advantage of the contemporary hardware.

The Transformer architecture consists of an encoder and a decoder. Transformer blocks consisting of a self-attention layer and a feed-forward neural network are stacked on top of one another where the output of one is passed as input to the next one. In the first layer, the words in the input text are converted to embeddings and positional encoding is added to these embeddings in order to add information about the word’s position. The word embeddings generated from the first block are passed to the next block as input. The final encoder generates an embedding for each word in the text. The original transformer architecture consists of a decoder stack which is used for machine translation. However, that is not required for classification tasks as we are only interested in classifying the input text using the embeddings generated by the encoder stack. We used two transformer-based architectures to adapt to the classification task.

\subsubsection{BERT.}
BERT-base \cite{devlin2018bert} is a model that contains 12 transformer blocks, 12 self-attention heads, and a hidden size of 786. The input for BERT contains embeddings for a maximum of 512 words and it outputs a representation for this sequence. The first token of the sequence is always [CLS] which contains the special classification embedding and another special token [SEP] is used for separating segments for other NLP tasks. For the purpose of a classification task, the hidden state of the [CLS] token from the final encoder is considered and a simple softmax classifier is added on top to classify the representation.

\subsubsection{DistilBERT.}
DistilBERT \cite{sanh2019distilbert} offers a simpler, cheaper, and lighter solution that has the basic transformer architecture similar to that of BERT. Instead of distillation during the fine-tuning phase specific to the task, here the distillation is done during the pre-training phase itself. The number of layers is halved and algebraic operations are optimized. Using a few such changes, DistilBERT provides competitive results even though it is 40\% smaller than BERT.

\section{Experimental setup }
\subsection{ Dataset details}
The Contraint@AAAI 2021 Covid-19 Fake news detection dataset\cite{patwa2020fighting} consists of tweets and their corresponding label. The label categorizes tweets as either fake or real. The dataset has a predefined train, test, and validation split. The train data has 6420 samples, test data has 2140 samples and validation data has 2140 samples; making it a total of 10,700 media articles and posts acquired from multiple platforms. Train data contains 3060 fake samples and 3360 real samples while validation and test data contain 1020 fake samples and 1120 real samples each. The fake tweets were collected from fact-checking websites like Politifact, NewsChecker, Boomlive \cite{PolitiFa30:online,HomeNews39:online,BOOMCoro43:online}, and from tools like Google fact-check-explorer and IFCN chatbot \cite{FactChec32:online}. For obtaining real tweets verified Twitter handles were used.

After performing the pre-processing steps mentioned in section \ref{preprocessing} statistics of the dataset are shown in Table \ref{statistics}.  It is also observed that 2998 unique tokens from the test data are absent in the training dataset. Similarly, for the validation dataset, 2888 tokens are absent in the train dataset.

\begin{table}
\caption{Statistics of the dataset.}\label{statistics}
\begin{center}
\begin{tabular}{|l|l|l|l|}
\hline
Feature &  Train data & Test data & Validation data\\
\hline
Total words & 115244 & 39056 & 38021\\
Total unique tokens &  14264 & 7151 & 6927\\
Maximum length of a tweet (in words) & 871 & 968 & 209\\
Average length of a tweet (in words) & 17.95 & 18.25 & 17.76\\
\hline
\end{tabular}
\end{center}

\end{table}
Models like BERT are trained on huge text datasets like Wikipedia which comprise of text from a variety of domains. However, re-training such models on the corpus related to the domain under consideration might make the model adapt to a specific domain better. With this aim, an unlabelled corpus of covid tweets with the hashtag covid19 was gathered using Twitter API \cite{COVID19T32:online}. This corpus was used for further pretraining in BERT and Fast-Text related experiments reported in this paper.

\subsection{ Preprocessing of the dataset}\label{preprocessing}
Following steps of preprocessing are used for sequential models :

\begin{itemize}
    \item \textbf{Removal of HTML tags}: Often in the process of gathering dataset, web or screen scraping leads to the inclusion of HTML tags in the text. These tags are often not paid heed to but it is necessary to get rid of them.
    \item \textbf{Convert Accented Characters to ASCII characters}: To avoid the NLP model from treating accented words like "r\'esum\'e", "latt\'e", etc different from their standard spellings, the text has to be passed through this step. 
    \item \textbf{Expand Contractions}: Apostrophe is commonly used to shorten the entire word or a group of words. For example, "don't means "do not" and "it's" stands for "it is". These shortened forms are expanded in this step.
    \item \textbf{Removal of Special Characters}: Special characters are not readable because they are neither alphabets nor numbers. They include characters like "*", "\&", "\$", etc.
    \item \textbf{Noise Removal}: Noisy text includes unnecessary new lines, white spaces, etc. Filtering of such text is done in this process.
    \item \textbf{Normalization}: The entire text is converted into lowercase characters due to the case sensitive nature of NLP libraries.
    \item \textbf{Removal of stop-words}: English language stop words include words like ‘a’, ‘an’, ‘the’, ‘of’, ‘is’, etc which commonly occur in sentences and usually add less value to the overall meaning of the sentence. To ensure less processing time it is better to remove these stop words and let the model focus on the words that convey the main focus of the sentence.
    \item \textbf{Stemming}: This step reduces the word to its root word after removing the suffixes. But it does not ensure that the resulting word is meaningful. Among many available stemming algorithms, the one used for this paper is Porter’s Stemmer algorithm.
\end{itemize}

Sequential models were trained using two types of word embeddings namely Glove and Fast-text.
\begin{itemize}
    \item 100 dimensional pre-trained Glove \cite{pennington2014glove} embeddings
    \item 300 dimensional Fast-text \cite{joulin2016bag} embeddings which were generated by training on a joint corpus of train data, validation data specific to this task and covid19 corpus \cite{COVID19T32:online} of tweets.
\end{itemize}
The embedding layer is kept trainable and connected to the first layer of the respective network.


\subsection{Training Details}
All the models were trained using the Tensorflow 2.0 framework. All models were trained for a maximum of 10 epochs and validation loss was used to pick the best epoch.
\subsubsection{Transformer-based architectures.}
The transformer-based models BERT and DistilBERT are used in two different ways:

\paragraph{Fine-tuning strategies:}  BERT and DistilBERT models which are pre-trained on a general corpus can be used for different classification and generation tasks. We have fine-tuned these two models in order to adapt to the target classification task. Along with this, we have also used two publicly shared BERT-based models pretrained on covid corpus from the huggingface model hub. 
\begin{itemize}
    
\item Covid-bert-base : Covid-bert-base \cite{deepsetc5:online} is a pretrained model from huggingface which is trained on a covid-19 corpus using the BERT architecture.
\\
\item Covid-Twitter-Bert : Covid-Twitter-Bert \cite{muller2020covid} is pretrained using a large corpus of covid-19 twitter messages on BERT architecture. This model is used from huggingface pretrained models \cite{wolf2020transformers} and fine-tuned on the target dataset.

\end{itemize}

\paragraph{Further pretraining:} The pre-trained models of BERT and DistilBERT are based on a general domain corpus from the pre-covid era. They can be further trained on a corpus related to the domain of interest. In this case, we used an accumulated collection of tweets\cite{COVID19T32:online} with the hashtag covid19. These models were trained as a language model on the corpus of COVID-19 tweets which is also the target domain. This pre-trained language model was then used as a classification model in order to adapt to the target task. We manually pre-trained BERT and DistilBert models on a covid tweets dataset using huggingface library.

\section{Results and Discussion}
\begin{table}[]
\caption{Results using five strategies}\label{results}
\begin{center}
\begin{tabular}{cccc}
\hline
\textbf{Strategies} & \textbf{Model} & \multicolumn{2}{c}{\textbf{Accuracy}} \\
\hline
                    &               &  \textbf{Validation} & \textbf{Testing}\\
\hline
\multirow{1}{*}{Finetuning} & bert-cased & 97.94 & 98.08 \\
                     & distilbert-cased &  97.94 & 97.75\\
                     & bert-uncased & \bfseries98.13 & 97.71    \\
                     & distilbert-uncased & 97.94 & 98.22   \\
                     & covid-base-bert & 97.05 & 97.05   \\
                     & covid-twitter-bert &  98.22 & \bfseries98.36  \\
\hline
\multirow{1}{*}{LM Pretraining}& bert-cased & 98.04 & \bfseries98.41\\                                 & distilbert-cased & 98.13 &98.22 \\
                                & distilbert-uncased & 97.99 & 98.04\\
                            & bert-uncased & \bfseries98.27 & 98.17\\
\hline
\multirow{1}{*}{Fast-text} & CNN & 91.64 & 94    \\
                    & LSTM &  93.6 & 94.95  \\
                    & BiLSTM + Attention & 92.71 & 94.71    \\
                    & HAN & \bfseries95.42 & \bfseries95    \\
\hline
\multirow{1}{*}{GloVe} & CNN & 93 & 93.50\\
                 & LSTM & 92.52 & 92.62\\
                 & BiLSTM + Attention & \bfseries94.39 & 92.99\\
                 & HAN & 94.16 & \bfseries94.25\\
\hline
Baseline & -- & 93.46 & 93.32\\
\hline
\end{tabular}
\end{center}
\end{table}

We analyze the accuracies reported using different types of models on the target dataset in Table \ref{results}. The baseline accuracy refers to the best accuracy reported in \cite{patwa2020fighting} using SVM model. The BERT and DistilBERT models pretrained on the Covid-19 tweets corpus perform better than the ones which are only fine-tuned on the dataset. The bert-cased model which was trained manually on the covid-19 tweets corpus gives the best results followed by the Covid-Twitter-Bert model. Among the non-transformer models, HAN gives the best results. Overall, the transformer models both pre-trained and fine-tuned, perform much better than the non-transformer models word-based models. The fast text word vectors were trained on target corpus and hence perform slightly better than pre-trained GloVe embeddings. This shows the importance of pre-training on target domain like corpus.

\section{Conclusion}
Under the shared task of Contraint@AAAI 2021 Covid-19 Fake news detection, we analyzed the efficacy of various deep learning models. We performed thorough experiments on transformer-based models and sequential models. Our experiments involved further pretraining using a covid-19 corpus and fine-tuning the transformer-based models. We show that manually pretraining the model on a subject-related corpus and then adapting the model to the specific task gives the best accuracy. The transformer-based models outperform other basic models with an absolute difference of 3-4\% in accuracy. We achieved a maximum accuracy of 98.41\% using language model pretraining on BERT over the baseline accuracy of 93.32\%. Primarily we demonstrate the importance of pre-training on target domain like corpus.

\section*{Acknowledgements}
This research was conducted under the guidance of L3Cube, Pune. We would like to express our gratitude towards our mentors at L3Cube for their continuous support and encouragement. We would also like to thank the competition organizers for providing us an opportunity to explore the domain.


\bibliographystyle{splncs03}
\bibliography{main}

%







\end{document}